\documentclass[10pt,twocolumn,letterpaper]{article}
\usepackage{iccv}
\usepackage{times}
\usepackage{epsfig}
\usepackage{graphicx}
\usepackage{amsmath}
\usepackage{amssymb}
\usepackage[breaklinks=true,bookmarks=false]{hyperref}

\iccvfinalcopy % *** Uncomment this line for the final submission

\def\iccvPaperID{20} % *** Enter the ICCV Paper ID here

\ificcvfinal\fi

\begin{document}

%%%%%%%%% TITLE
\title{A Multi-task Mean Teacher for Semi-supervised Facial Affective Behavior Analysis}

\author{Lingfeng Wang$^{*1}$, Shisen Wang$^{*1}$, Jin Qi$^1$, Kenji Suzuki$^2$\\
	$^1$University of Electronic Science and Technology of China, Chengdu, China\\
	$^2$Tokyo Institute of Technology, Tokyo, Japan\\
	{\tt\small \{wanglingfeng,shisenwang\}@std.uestc.edu.cn, jqi@uestc.edu.cn, suzuki.k.di@m.titech.ac.jp}
}

\maketitle
\ificcvfinal\thispagestyle{empty}\fi
\let\thefootnote\relax\footnotetext{$^*$Equal Contribution.}

%%%%%%%%% ABSTRACT
\begin{abstract}
   Affective Behavior Analysis is an important part in human-computer interaction. Existing multi-task affective behavior recognition methods suffer from the problem of incomplete labeled datasets. To tackle this problem, this paper presents a semi-supervised model with a mean teacher framework to leverage additional unlabeled data. To be specific, a multi-task model is proposed to learn three different kinds of facial affective representations simultaneously. After that, the proposed model is assigned to be student and teacher networks. When training with unlabeled data, the teacher network is employed to predict pseudo labels for student network training, which allows it to learn from unlabeled data. Experimental results showed that our proposed method achieved much better performance than baseline model and ranked 4th in both competition track 1 and track 2, and 6th in track 3, which verifies that the proposed network can effectively learn from incomplete datasets.
\end{abstract}

%%%%%%%%% BODY TEXT
\section{Introduction}

Facial affective behavior recognition plays an important role in human-computer interaction \cite{kollias2021analysing}. It allows computer systems to understand human feelings and behaviors, which makes human computer interaction more applicable. Existing research used different approaches to represent human emotions, such as valence-arousal estimation (VA), facial action unit (AU) detection, and facial expression (Expr) classification.

In the challenges of the 2nd Affective Behavior Analysis in-the-wild (ABAW2) Competition \cite{kollias2021analysing, kollias2020analysing,kollias2021distribution,kollias2021affect,kollias2019expression,kollias2019face,kollias2019deep,zafeiriou2017aff}, the organizers collect a large scale in-the-wild database Aff-Wild2 to provide a benchmark for the three emotion representation tasks. There are strong correlations between the three different tasks. Multi-task learning can extract joint features from the correlated tasks and provide better performance than training on a single task. In the last year’s competition, some teams proposed multi-task learning model to explore the learning of multiple correlated tasks simultaneously. For example, Two-Stream Aural-Visual model (TSAV) \cite{Kuhnke_2020} achieved superior performance in a multi-task manner.

 However, the labels of ABAW2 Competition database are incomplete. Even through the three tasks share the same video database, most videos in the dataset are only labeled for one or two tasks. During the multi-task training process, only tasks with labels can be trained while other tasks without labels are ignored, which is quite inefficient. Previous studies also faced this challenge and had to  treat different tasks independently. Those methods can only make use of limited labeled data while ignoring abundant incomplete labeled data. Hence, it is highly desirable to leverage additional unlabeled data to improve the performance.
 
To tackle this problem, we develop a multi-task mean teacher \cite{tarvainen2017mean} framework to boost affective behavior recognition performance in a semi-supervised manner. We firstly proposed an audio-video model to learn the three tasks mutually.  Our model shares the same backbone with TSAV while differing in a preprocessing step and output layers. Second, we take this multi-task model as both student network and teacher network. For labeled tasks, the supervised losses on all labeled tasks are integrated as multi-task supervised loss. For unlabeled tasks, we enforce the outputs of the student network and the teacher network to be consistent with each other using consistency loss. By adding the supervised loss and the consistency loss together, our network can be trained with both labeled and unlabeled data. Howerver, prediction results of the teacher network could be inconsistent or incorrect, which is harmful to model training. To address this problem, we employ self-attention importance weighting, ranking regularization modules described in self-cure network \cite{wang2020suppressing}, to suppress the impact of uncertainties and to prevent deep networks from over-fitting uncertain facial expression. 

With these improvements, our model achieves a competitive result in the competition. At the second ABAW competition, Our proposed method ranked fourth for both valence-arousal estimation and expression classification tasks. It is worth mentioning that our model was trained on the competition database only but achieved comparable performance with large scale pretrained models with extra datasets proposed by other teams. 

Our major contributions are summarized as: 

• First, we developed a multi-task multi-modal model for simultaneously analyzing valence-arousal estimation, facial action unit detection, and expression classification.

• Second, we designed a mean teacher framework to fuse consistency loss of incomplete labeled data with suoervised loss from labeled data. We also made a step forward to adopt self-attention importance weighting, ranking regularization to solve the uncertainty problem of pseudo labels. In this way, our proposed model can effectively leverage both labeled and unlabeled data. 

%------------------------------------------------------------------------
\section{Related works}

Previous studies on the Aff-Wild2 have proposed some effective facial affective behavior analysis models, especially to explore inter-task correlations using multi-task learning. In \cite{kollias2021distribution}, Kollias \etal proposed FaceBehaviorNet for large-scale face analysis, by jointly learning multiple facial affective behavior tasks and a distribution matching approach. Kuhnke \etal~\cite{Kuhnke_2020} proposed a two-stream aural-visual network to combine vision and audio information for multi-task emotion recognition. 

Even though the mentioned multi-task methods obtained promising results, they did not solve incomplete label problem of Aff-Wild2 dataset. Deng \etal~\cite{deng2020multitask} proposed a data-driven teacher model to fill in the missing labels. They trained a teacher model firstly. After that, the distillation knowledge technique is applied to train a student model. Inspired by their work, in this paper we propose a mean teacher model to leverage incomplete labeled data. Different from Deng’s work, the teacher model and student model in our framework are trained at the same time. Such an end-to-end training process is more flexible than Deng’s method.

%------------------------------------------------------------------------
\section{Methodology}
\subsection{Multi-task Affective Behavior Recognition Model}
\begin{figure}
	\begin{center}
		\includegraphics[width=1.0\linewidth]{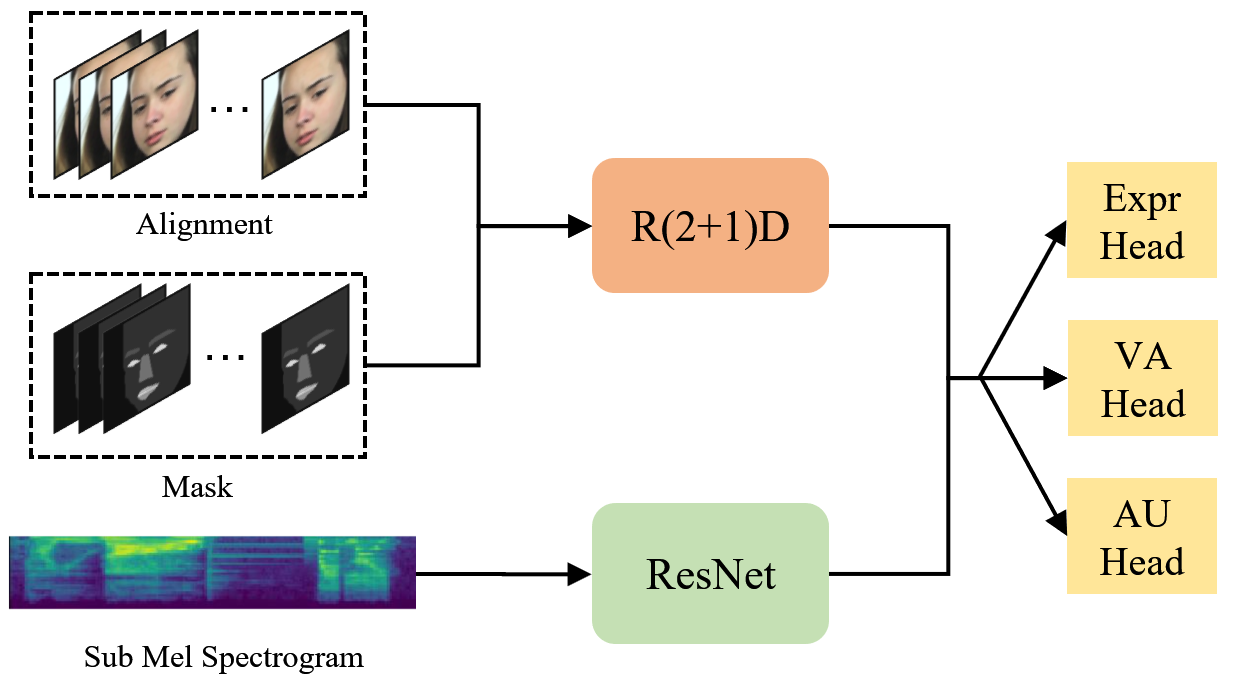}
	\end{center}
	\caption{Framework of multi-task affective behavior analysis model.}
	\label{fig:model}
\end{figure}

Figure~\ref{fig:model} shows the framework of our multi-modal affective behavior analysis model. The audio-video dual branch architecture is inspired by TSAV. The multi-modal model fuse features of two branches to give prediction on three different emotion representation tasks.

\begin{figure}[t]
	\begin{center}
		\includegraphics[width=0.7\linewidth]{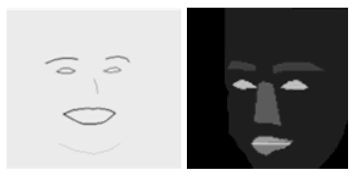}
	\end{center}
	\caption{Left: Mask of TSAV; Right: Proposed mask.}
	\label{fig:mask}
\end{figure}

For the Visual stream, the input clips are composed of cropped aligned images and corresponding face masks. The usage of the face mask in TSAV is believed to be the most helpful for its performance. We use HRNet \cite{cheng2020higherhrnet} to detect 106 facial landmarks and render a face segmentation mask for every face image. As shown in Figure~\ref{fig:mask}, comparing to the mask rendering method of TSAV, which can only render contours image using 68 landmarks, our method can provide more semantic information. 

After preprocessing, each frame image has 4 channels (RGB + mask). These frames are sampled from video with dilation of 6, and constitute input clips with 8 frames. We employ pre-trained (R2+1)D \cite{tran2018closer} as a visual model to extract spatio-temporal information from the visual stream.

As for audio stream, we followed the setting of TSAV. We compute a mel spectrogram for all audio stream extracted from the video using TorchAudio \cite{paszke2019pytorch} package. For each clip, a spectrogram is cut into a smaller sub-spectrogram with the center of sub-spectrogram aligning with the current frame at time $t$. A ResNet-18 \cite{he2016deep} is used for mel spectrogram analysis.

Finally, the output features extracted by the video branch and audio branch are merged. Prediction heads share the same fused features and give final predictions on the three expression representation tasks. AU and VA heads are fully connected layers that map features to AU classification and valence-arousal regression respectively. Expr head contains fully connected layers with self-attention module for expression classification.

\subsection{Mean Teacher}
\begin{figure*}
	\begin{center}
		\includegraphics[width=0.8\linewidth]{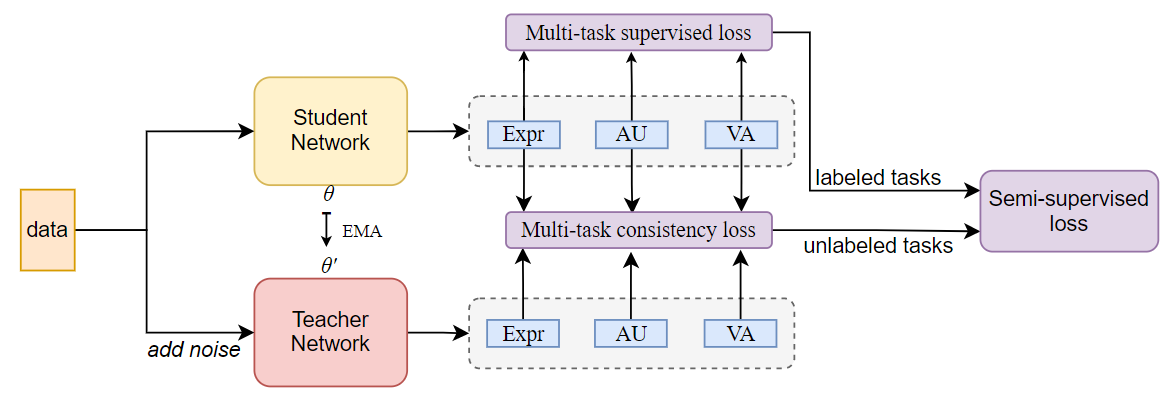}
	\end{center}
	\caption{Framework for Mean Teacher.}
	\label{fig:mean_teacher}
\end{figure*}

In the Aff-Wild2 dataset, only 144 videos contain labels for all the three different tasks. The rest 297 videos only have labels for one or two tasks, which brought a challenge for multi-task model training. Specifically, different tasks cannot be supervised at the same time during the multi-task training process. There are different numbers of labeled data for the three tasks, which could lead to imbalanced performance among different tasks.  In order to train three tasks at the same time, we introduced the mean teacher \cite{tarvainen2017mean} to take advantage of semi-supervised learning, as shown in Figure~\ref{fig:mean_teacher}.

The mean teacher framework is extended from a supervised architecture by making a copy of the original model. The original model is called a student and the new one is called the teacher. The parameters of a teacher network are updated by computing the exponential moving average (EMA) of student model’s parameters. Updating parameters of the teacher model and student model by interleaving can reduce overfitting due to the additional unlabeled data.

At each mini-batch in training process, the same batch data is the input to both the student and the teacher. Random noise is added to input data of the teacher to enforce the model to keep consistency under random disturbances. Here we apply random brightness augmentation for each input clip of the teacher model.

For tasks with labels, we can calculate the supervision loss with ground truth. For the unlabeled tasks, we take the predictions of the teacher network as a hard label and then enforce the predictions of student network to be consistent with the hard label. In this way, the optimizer can update the weights of the student network normally for both labeled and unlabeled tasks.
After each training step, the weights of the teacher network are updated by calculating the exponential moving average (EMA) of the student weights, which can be understood as ensemble of student models. At the $t$ training iteration, the parameters of the teacher network are
\begin{equation}
	\theta_{t}^{\prime}=\eta \theta_{t-1}^{\prime}+(1-\eta) \theta_{t}
\end{equation}
where $\theta_{t}$ represents the parameters of model, $\eta$ is a hyper parameter of moving average. Here we choose $\eta=0.99$, as suggested in \cite{tarvainen2017mean}.

\subsection{Self-Cure module for uncertainty suppression}
Comparing to ground truth labels, pseudo labels predicted by the teacher model could be inconsistent or incorrect. Training with these uncertain labels may cause model overfitting to incorrect samples, especially in the expression recognition task. We introduce self-cure module described in self-cure network \cite{wang2020suppressing} to solve this problem in expression classification head. The key idea of the self-cure module is to allow network to learn to decide which labels are correct. It consists of two parts: a self-attention importance weighting module and a rank regularization module.

The self-attention importance weighting module can learn the importance of each sample by predicting an attention weight.
Attention weights are predicted by a fully connected layer with sigmoid functions. Samples with higher weights are more reliable and more important for training. For each input sample, its expression classification output is multiplied to the importance weight predicted by self-attention weighting module and give final prediction.

With predicted weights, the ranking regularization module ranks these weights and split them into two groups, which are a high-importance group and a low-importance group. The two groups are regularized by forcing a margin between the two groups with a rank regularization loss (RR-Loss):

\begin{equation}
	\mathcal{L}_{R R}=\max \left\{0, \delta-\left(\alpha_{H}-\alpha_{L}\right)\right\}
\end{equation}
where $\alpha_{H}$ is the mean weight of high-importance group, $\alpha_{L}$ is the mean weight of low-importance group. $\delta$ represents a margin, which is set to be $0.15$.

Ranking regularization module and self-attention weighting module together ensures model learns meaningful training data by highlighting certain samples and suppressing uncertain samples in our semi-supervised learning framework.

\subsection{Loss Function}

Loss function for our semi-supervised model consists of two parts: multi-task supervised loss and multi-task consistency loss. Supervised loss is computed with ground truth if label is available. When a label is missing, we take the prediction results of the teacher model as a hard label. Then we calculate consistency losses between the hard label and the prediction of student model in the same way as supervised.  
\begin{equation}
	L_{t}=\left\{\begin{array}{c}
		L_{t}^{s}=L_{t}(S, G), \text {\emph{label available}} \\
	L_{t}^{c}=L_{t}(S, T), \text {\emph{label missing}}
	\end{array}\right.
\end{equation}
where $L_{t}$ denotes a loss for task $t$, $L^{s}$ is a supervised loss, $L^{c}$ is a consistency loss; $S$ and $T$ represent the predictions of the student network and teacher network respectively, and $G$ is ground truth. 

For expression classification, we used the sum of the cross entropy and rank regularization loss as objective function:
\begin{equation}
L_{E x p r}=L_{C E}+L_{R R}
\end{equation}

For AU task, we define the total binary cross entropy loss by
\begin{equation}
	L_{A U}=-\sum_{i}^{12}\left\{y_{i} \cdot \log \left(o_{i}\right)+\left(1-y_{i}\right) \cdot \log \left(1-o_{i}\right)\right\}
\end{equation}
where $y$ is the 12 dimensional label vector, $o$ is the corresponding prediction vector.

%For valence and arousal task, we used the sum of the CCC for valence and the CCC for valence arousal as supervised loss:
The concordance correlation coefficient (CCC) loss \cite{kollias2019expression} is used for valence and arousal estimation:
\begin{equation}
		L_{V A}=\frac{1}{2} \times(C C C_{V}+C C C_{A})
\end{equation}

%For a categorical expression classification task, we use categorical cross entropy and ranking regularization loss together. The binary cross entropy is used for action unit detection and the concordance correlation coefficient (CCC) loss \cite{kollias2019expression} for valence and arousal estimation.
The final total loss for current batch is the sum of losses for expression, action unit, and valence and arousal estimation tasks, defined as follows:
\begin{equation}
	L_{total}=w_{1} L_{Expr}+w_{2} L_{AU}+w_{3} L_{VA}
\end{equation}
In this paper, we set $w{1} = 1.0$, $w{2} = w{3} = 0.3$.
%------------------------------------------------------------------------
\section{Experiments}
\subsection{Dataset}
The proposed model was trained on the large-scale in-the-wild Aff-Wild2 dataset only. This dataset contains 564 videos with frame-level annotations for valence-arousal estimation, facial action unit detection, and expression classification tasks. We use the official provided cropped and aligned images in the Aff-wild2 dataset. Additionally, we rendered corresponding facial masks for all the cropped images as described in Section 3.1.

We split the training and validation set by ourselves instead of using official validation set. The official database do not have the same training and validation split among the three tasks. For example, some videos belonging to validation set of AU task also appear in training set of expression classification and valence-arousal estimation tasks. When validating AU task using the inconsistent split data, prior knowledge from training other task will affect the evaluation. Thus, we create a custom training and validation split to ensure that three tasks share consistent split. Moreover, we keep the samples in each task to be split into training and validation set at a ratio of 8:2.

\subsection{Training Setup}
A model was trained with our training split dataset only. We used the pretrained weight from TSAV to initialize the backbone for audio and video branch. The model was optimized using Adam optimizer and a learning rate of 0.0005. Random brightness augmentation was applied for each input clip. The mini-batch size was set to 32. The training and validating processes were performed using two GPU to allocate each of the teacher and student networks to one GPU. 

\subsection{Results}
We used the same metrics as suggested in \cite{kollias2021analysing} to evaluate performance.
The metrics of the three tasks are defined as follows:
\begin{equation}
		M_{VA}=\frac{1}{2} \times(C C C_{V} + C C C_{A})
\end{equation}
\begin{equation}
		M_{Expr}=0.67 \times F_{1}+0.33 \times A c c
\end{equation}
\begin{equation}
		M_{AU}=\frac{1}{2} \times (F_{1} + A c c)
\end{equation}

\begin{table}
	\begin{center}
		\begin{tabular}{|c|ccc|}
			\hline
			%Method & \tabincell{c}{$M_{Expr}$} & \tabincell{c}{$M_{VA}$} & \tabincell{c}{$M_{AU}$} \\
			Method & $M_{Expr}$ & $M_{VA}$ & $M_{AU}$ \\
			\hline\hline
			Basic model & 0.475 & 0.513 & 0.623 \\
			Basic model+MT  & 0.489 & 0.566 & 0.674\\
			Basic model+SC+MT &	\textbf{0.501} & \textbf{0.568} & \textbf{0.675}\\
			\hline
		\end{tabular}
	\end{center}
	\caption{Performance of our models on validation set.  Expr, VA, and AU mean the score for each task. MT denotes the mean teacher, SC denotes the self-cure module.}
	\label{table:evaluation}
\end{table}

\begin{table*}
	\begin{center}
		\resizebox{\textwidth}{!}{
		\begin{tabular}{|c|ccc|ccc|ccc|}
			\hline
			%Method & \tabincell{c}{$F_{1}$(Expr)} & \tabincell{c}{Acc(Expr)} & \tabincell{c}{$M_{Expr}$} & \tabincell{c}{$CCC_{V}$} &\tabincell{c}{$CCC_{A}$} & \tabincell{c}{$M_{VA}$}&\tabincell{c}{$F_{1}$(AU)}&\tabincell{c}{Acc(AU)}&\tabincell{c}{$M_{AU}$} \\
			Method & $F_{1}$(Expr) & Acc(Expr) & $M_{Expr}$ & $CCC_{V}$ &$CCC_{A}$ & $M_{VA}$&$F_{1}$(AU)&Acc(AU)&$M_{AU}$ \\
			\hline\hline
			Baseline \cite{kollias2021analysing} & 0.260 & 0.460 & 0.326 & 0.200 & 0.190 & 0.195 & 0.367 & 0.193 & 0.280\\
			NISL-2021 & 0.431 & 0.654 & 0.505& \textbf{0.533} & 0.454 & \textbf{0.494} & 0.451 & 0.847 & 0.653 \\
			Netease Fuxi Virtual Human\cite{zhang2021prior}  & \textbf{0.763} & \textbf{0.807} & \textbf{0.778} & 0.486 & 0.495 & 0.491 & \textbf{0.506} & \textbf{0.888} & \textbf{0.697}\\
			Our &	0.476 & 0.732 & 0.560& 0.478 & \textbf{0.498} & 0.488 & 0.394 & 0.875 & 0.634\\
			\hline
		\end{tabular}}
	\end{center}
	\caption{Results on the test set of the Aff-Wild2 dataset. The best result is indicated in bold}
	\label{table:test}
\end{table*}

In order to analyze the effects of our proposed framework design, we conducted ablation studies to compare performance with or without proposed components. The results on the validation set can be seen in Table \ref{table:evaluation}. Note that the basic multi-task model is trained with complete labeled data only, while the model with a mean teacher (MT) framework is trained with incomplete labeled data. The usage of the mean teacher allows the model to learn from the unlabeled data which incurs a superior affective behavior recognition performance on each task. However, performance improvement in expression task is not as significant as the other two tasks. We made further investigation by using self-cure mechanism to suppress the impact of uncertainties and further improve expression recognition performance. Experiment indicates that the self-cure module resolves the problem and achieves best performance on three benchmarks.

We also evaluated our model on the official test set. The results on the test set can be seen in Table \ref{table:test}. Our model outperforms the baseline model of \cite{kollias2021analysing} a lot. As for VA track, our model achieve comparable performance comparing to model of team NISL-2021, especially in CCC Arousal metric.As for Expr and AU tracks, our accuracy is comparable in contrast to model of \cite{zhang2021prior}, which verifies that the proposed network can effectively learn from incomplete datasets. But our $F_{1}$ score is much lower. The first reason is that they use large scale pre-trained model on additional dataset, whereas we only use the Aff-Wild2 database for training. Another reason is that we did not use data balancing strategy, which lead to poor $F_{1}$ score. We will investigate data balancing strategies in future research.
%------------------------------------------------------------------------
\section{Conclusions}
This paper presents a semi-supervised facial affective behavior recognition model by developing a multi-task mean teacher framework. Our key idea is to firstly develop a multi-modal model to recognize the three emotion representation tasks. Then we employ the mean teacher with semi-supervised learning to learn from additional unlabeled data for further improving the recognition performance. Experimental results on validation datasets show that our semi-supervised model outperforms the original supervised model in all tasks, which verifies the effectiveness of the proposed method. For future work, we plan to resolve the problem of data imbalance to further boost the performance.
{\small
	\bibliographystyle{unsrt}
	\bibliography{egbib}
}

\end{document}